\documentclass{article}
\usepackage{arxiv}

\usepackage[pdftex]{graphicx}  
\usepackage[utf8]{inputenc} 
\usepackage[T1]{fontenc}    
\usepackage{hyperref}       
\usepackage{url}            
\usepackage{booktabs}       
\usepackage{amsfonts}       
\usepackage{nicefrac}       
\usepackage{microtype}      
\usepackage{lipsum}
\usepackage{amsmath}
\usepackage[numbers, sort&compress]{natbib}

\DeclareUnicodeCharacter{FB01}{fi}

\title{Personalized Activity Recognition with Deep Triplet Embeddings}

\author{
  David M.~Burns \\
  $^1$Division of Orthopaedic Surgery \\
  $^2$Sunnybrook Research Institute \\
  University of Toronto \\ 
  Toronto, Ontario, Canada \\
  \texttt{d.burns@utoronto.ca} \\
  \And
  Cari M.~Whyne \\
  $^1$Division of Orthopaedic Surgery \\
  $^2$Sunnybrook Research Institute \\
  University of Toronto \\
  Toronto, Ontario, Canada \\
  \texttt{cari.whyne@sunnybrook.ca} \\
}

\begin{document}
\maketitle

\begin{abstract}
A significant challenge for a supervised learning approach to inertial human activity recognition is the heterogeneity of data between individual users, resulting in very poor performance of impersonal algorithms for some subjects. We present an approach to personalized activity recognition based on deep embeddings derived from a fully convolutional neural network. We experiment with both categorical cross entropy loss and triplet loss for training the embedding, and describe a novel triplet loss function based on subject triplets. We evaluate these methods on three publicly available inertial human activity recognition data sets (MHEALTH, WISDM, and SPAR) comparing classification accuracy, out-of-distribution activity detection, and embedding generalization to new activities. The novel subject triplet loss provides the best performance overall, and all personalized deep embeddings out-perform our baseline personalized engineered feature embedding and an impersonal fully convolutional neural network classifier. 
\end{abstract}


\section{Introduction}
Inertial sensors embedded in mobile phones and wearable devices are commonly employed to classify and characterize human behaviors in a number of applications including  tracking fitness, elder safety, sleep, physiotherapy, and others \cite{sousa_lima_human_2019, mertz_convergence_2016, metcalf_wearables_2016, piwek_rise_2016}. Improving the accuracy and robustness of the algorithms underlying inertial human activity recognition (HAR) systems is an active field of research. 

\paragraph{}
A significant challenge for a supervised learning approach to inertial human activity recognition is the heterogeneity of data between individual users. This heterogeneity occurs in relation to heterogeneity in the hardware on which the inertial data is collected \cite{stisen_smart_2015, qi_examining_2018}, different inherent capabilities or attributes relating to the users themselves \cite{modave_mobile_2017}, and differences in the environment in which the data is collected \cite{robert-lachaine_effect_2017}. 

\paragraph{}
Large data sets incorporating the full spectrum of user, device, and environment heterogeneity may represent a potential approach to addressing these challenges, however, this approach presents significant logistical and financial challenges. Further, the devices and sensors on which inertial data is collected continuously evolve over time. It may not be feasible to train generic supervised algorithms that perform equally well for all users and devices. An alternative is to leverage labeled user-specific data for a personalized approach to activity recognition. 

\subsection{Related Work}
Human activity recognition from inertial time series data has classically been conducted using a supervised learning approach with non-neural classifiers, after transformation of the data using an engineered feature representation consisting of statistical, time-domain, and or frequency-domain transforms \cite{bulling_tutorial_2014}. Modern supervised learning approaches using convolutional and or recurrent neural networks are increasingly utilized and have demonstrated improvements in classification accuracy over non-neural models \cite{sousa_lima_human_2019}. Both non-neural and neural network supervised learning models have been applied to personalized activity recognition \cite{weiss_impact_2012, sztyler_online_2017, zhao_user-adaptive_2018, meng_towards_2017, rokni_personalized_2018, cvetkovic_semi-supervised_2011, hong_toward_2016}. 

\paragraph{}
User-specific supervised learning models can be trained through one of three general schemes. First, a user-specific model can be trained de novo with user-specific data or a combination of generic and user-specific data. This is generally not feasible for neural network approaches that require vast data sets and computational resources for training, but works well for non-neural approaches with engineered features \cite{weiss_impact_2012}. Second, model updating (online learning, transfer learning) with user-specific data is feasible for both non-neural \cite{sztyler_online_2017, zhao_user-adaptive_2018, meng_towards_2017} and neural network supervised learning algorithms \cite{rokni_personalized_2018}. \citet{rokni_personalized_2018} trained a generic convolution neural network architecture and adapated it to specific users by retraining the classification layer while fixing the weights of the convolutional layers with excellent results. The third scheme involves using classifier ensembles \cite{cvetkovic_semi-supervised_2011, hong_toward_2016}. \citet{hong_toward_2016} trained non-neural models on subpopulations within the training set, and selected user-specific classifier ensembles based on testing the pre-trained classifiers on the user-specific data. These personalized methods have all produced favorable results in comparison to generic models. However, generating, validating, and maintaining user-specific supervised learning models presents its own logistical challenges in a production environment. There are also currently regulatory barriers to such an approach in the context of software as a medical device \cite{u.s._food_&_drug_administration_proposed_2019}. 

\subsection{Personalized Features}

An alternative approach to personalized activity recognition is to store an embedding of labeled user-specific data. The embedding may then be searched to classify and or characterize new unlabeled data for the user. Further benefits of this approach include the capacity to incorporate novel activity classes without model re-training, and identify out-of-distribution (OOD) activity classes, supporting an open-set activity recognition framework \cite{geng_recent_2018}. Similar to the typical supervised learning approach, the embedding can be specified a priori with engineered features, and or be learned from the data. 

\paragraph{}
The penultimate feature layer of neural network classifiers in various domains have been shown to be useful for classification, and other tasks (e.g. visualization, clustering) \cite{sermanet_overfeat:_2014, sani_learning_2017}. \citet{sani_learning_2017} demonstrated that features extracted from a deep convolutional neural network are superior for generic activity recognition in comparison to engineered features with non-neural models. However, features extracted from deep neural networks are often treated as a side effect of the classifier training, rather than being explicitly sought. Metric learning methods, such as Siamese Neural Networks (SNN) \cite{bromley_signature_1993} and Triplet Neural Networks (TNN) \cite{hoffer_deep_2018} optimize an embedding directly for the desired task. To our knowledge, these methods have not been applied to personalized activity recognition. 

\subsection{Purpose}
In this manuscript we experiment with deep embeddings for personalized activity recognition, specifically considering 1) extracted features from a neural network classifier and 2) an optimized embedding learned using TNNs. We compare these to a baseline impersonal neural network classifier, and a personalized engineered feature representation. 

\section{Methods}

\subsection{Data Sets}

The algorithms were evaluated on three publicly available inertial activity recognition data sets: MHEALTH \cite{banos_design_2015}, WISDM \cite{weiss_smartphone_2019}, and SPAR \cite{burns_shoulder_2018}. These data sets encompass a combination of activities of daily living, exercise activity, and physiotherapy activities. Class balance is approximately equal within each and there is minimal missing data. The specific attributes of these data sets is summarized in Table \ref{psm:t:data}. 

\begin{table}[!ht]
\centering
\begin{tabular}{cccccc}
\toprule
\textbf{Data Set} & \textbf{Sensors}          & \textbf{Subjects} & \textbf{Classes} & \textbf{Sampling} & \textbf{Domain} \\ \midrule
MHEALTH \cite{banos_design_2015}         & 9-axis IMU x3, 2-lead ECG & 10                & 12               & 100 Hz            & Exercise        \\
WISDM \cite{weiss_smartphone_2019}           & 6-axis IMU x2             & 51                & 18               & 20 Hz             & ADL, Exercise  \\
SPAR \cite{burns_shoulder_2018}             & 6-axis IMU x1             & 40                & 7                & 50 Hz             & Physiotherapy \\ \bottomrule \\
\end{tabular}
\caption{Data set characteristics. The MHEALTH data was collected with three proprietary inertial sensors on the subjects’ right wrist, left leg, and chest. The WISDM data was collected from an Android smart watch worn by the subjects, and a mobile phone in the subjects’ pocket. The SPAR data was collected from 20 subjects (40 shoulders) using an Apple smart watch. IMU: Inertial Measurement Unit, ADL: Activity of Daily Living.}
\label{psm:t:data}
\end{table}

\subsection{Preprocessing}

The WISDM and MHEALTH data was resampled to 50 Hz, using cubic interpolation, to provide a consistent basis for evaluating model architecture. The time series data were then pre-processed with sliding window segmentation to produce fixed length segments of uniform activity class. A four second sliding window was utilized for the MHEALTH and SPAR data sets, and a ten second window was utilized for WISDM for consistency with previous evaluations \cite{jordao_human_2019, weiss_smartphone_2019, burns_shoulder_2018} . An overlap ratio of 0.8 was used in the sliding window segmentation as a data augmentation strategy. 

We utilized only the smart watch data from the WISDM data set, because the smart watch and mobile phone data were not synchronized during data collection. We also had to exclude four WISDM subjects from the evaluation due to errors in the data collection that resulted in absent or duplicated sensor readings (subjects 1637, 1638, 1639, and 1640). 

\subsection{Impersonal Fully Convolutional Network (FCN)}
\label{psm:sec:fcn}
\begin{figure}[!ht]
  \begin{center}
    \includegraphics[scale=0.7]{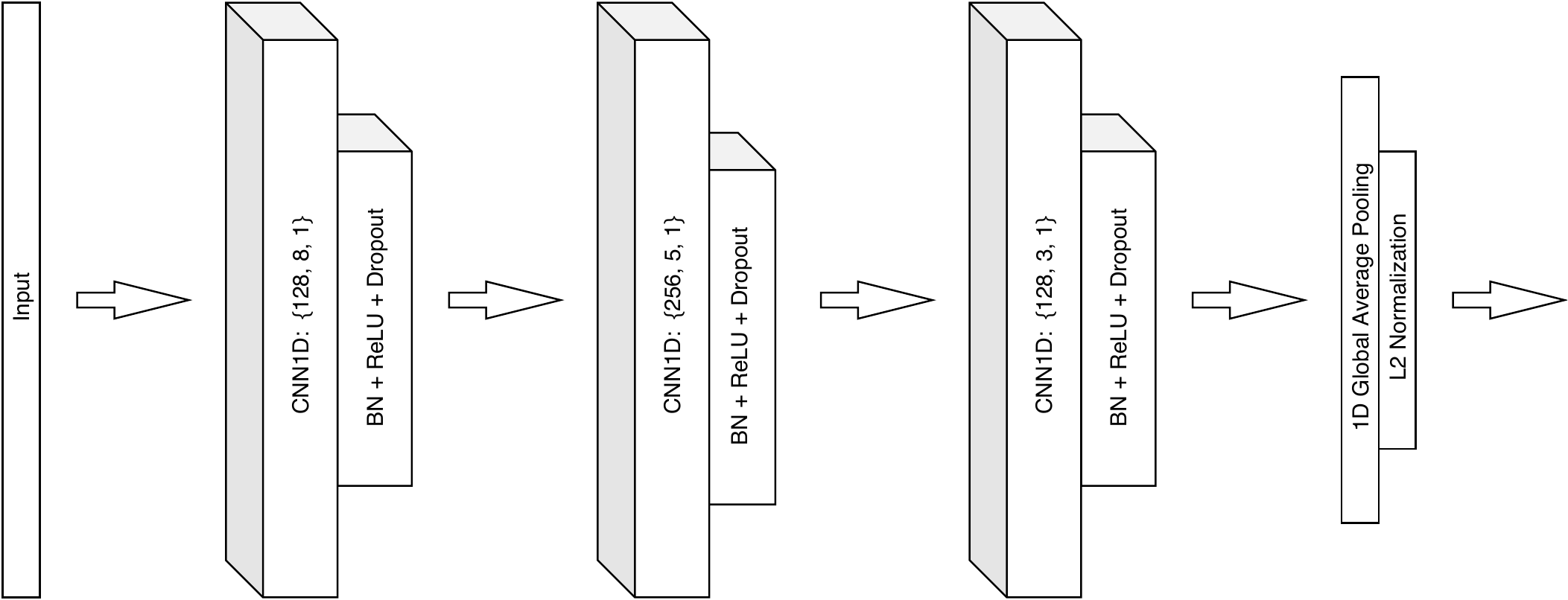}
  \end{center}
  \caption{Fully convolutional network (FCN) model core. 1D convolutional layers are defined by \{filters, kernel size, stride\}. A dropout ratio of 0.3 was used at each layer. The embedding size is equal to the number of filters in the last convolutional layer (128). The total number of parameters for this model is 270,848. An implementation is available for \texttt{keras} at  \texttt{https://github.com/dmbee/fcn-core} }
  \label{psm:f:fcn}
\end{figure}

A fully convolutional neural network (FCN) was utilized as the baseline impersonal supervised learning model. The FCN architecture is considered a strong baseline for time series classification even in comparison to deep learning models with modern architectural features used in computer vision such as skip connections \cite{wang_time_2016}. The FCN core (Figure \ref{psm:f:fcn}) consists of 1D convolutional layers, with rectified linear unit (ReLU) activation, and batch normalization. Regularization of the model is achieved using dropout applied at each layer. Global average pooling is used after the last convolutional layer to reduce the model sensitivity to translations along the temporal axis, as this ensures the receptive field of the features in the penultimate feature layer includes the entirety of the window segment. The receptive field of filters in the last convolutional layer prior to global average pooling was 13 samples, which is equivalent to 260 ms at a sampling rate of 50 Hz. An $L^2$ normalization is applied after global pooling to constrain the embedding to the surface of a unit hypersphere, which improves training stability.

\paragraph{}
The impersonal FCN classifier model consists of an FCN core with a final dense layer with softmax activation. The FCN classifier was trained for 150 epochs using the adam optimizer, categorical cross entropy loss, and a learning rate of 0.001. Gradient norm clipping to 1.0 was used to mitigate exploding gradients. 

\paragraph{}
The FCN core architecture was also used as the basis for both the personalized deep features (PDF) and personalized triplet network (PTN) models to provide a consistent comparison of these approaches.

\subsection{Personalized Feature Classifier} 
\begin{figure}[!ht]
  \begin{center}
    \includegraphics[scale=0.7]{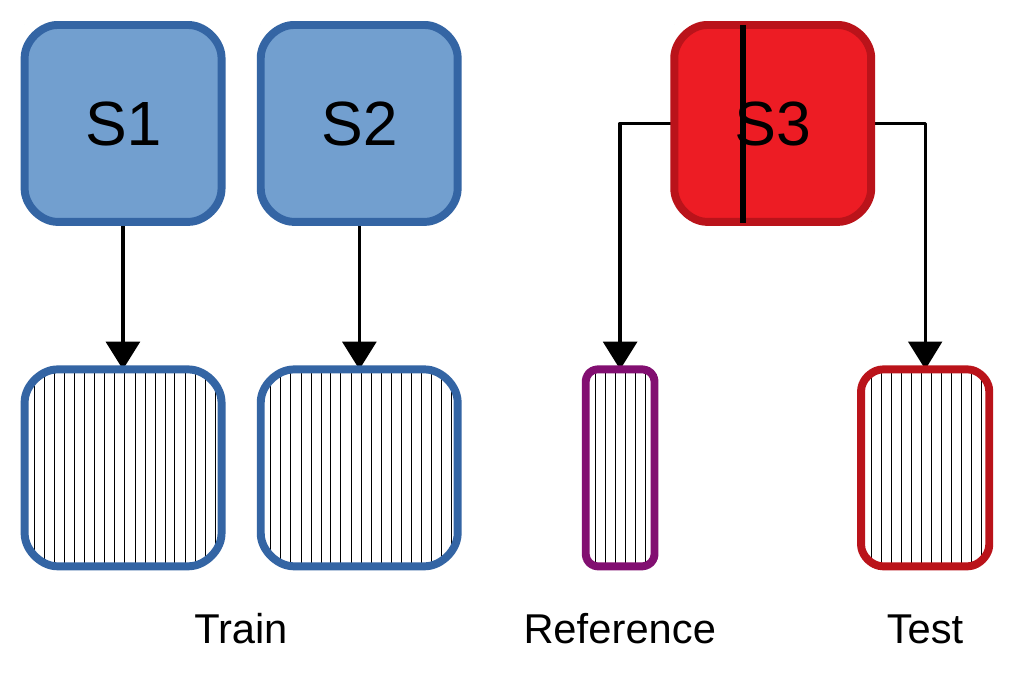}
  \end{center}
  \caption{Data splitting for evaluation of personalized feature classifiers. Folds are split by subject (S), with reference data derived from test subjects prior to sliding window segmentation ensuring no temporal overlap between train, reference, and test data.}
  \label{psm:f:datasplit}
\end{figure}

Three personalized feature classifiers: personalized engineered features (PEF), personalized deep features (PDF), and personalized triplet network (PTN) were implemented and compared for activity classification performance. Inference is achieved in each of these models by comparing a subject’s embedded test segments to labeled reference embeddings specific to the subject. For the test subjects, the time series data for each activity was split along the temporal axis, reserving the first part for reference data and the latter part for inference. This split was performed prior to sliding window segmentation to ensure there is no temporal overlap of reference and test samples. This partitioning of the data is depicted in Figure \ref{psm:f:datasplit}. We evaluate the effect of reference data size on model performance, using 50\% of the test data as the baseline evaluation. 

\paragraph{}
To determine the activity class in a test segment, we searched the reference embeddings for the 3 k-nearest neighbors (k-NN) using a euclidean distance metric and a uniform weight decision function. 

\subsubsection{Personalized Engineered Features (PEF)}
We used an engineered feature representation to serve as a baseline personalized classifier model. The representation consisted of typical statistical and heuristic features used for inertial activity recognition \cite{gonzalez_features_2015} including mean, median, absolute energy, standard deviation, variance, minimum, maximum, skewness, kurtosis, mean spectral energy, and mean crossings. The features were individually computed for each of the data channels in the data set. This resulted in 66 features for the WISDM and SPARS data sets, and 174 features for the MHEALTH data set. All features were individually scaled to unit norm and zero mean across the training data set. 

\subsubsection{Personalized Deep Features (PDF)}
The personalized deep feature model is identical to the FCN model in architecture and training. However, inference is achieved by embedding reference samples with the FCN penultimate feature layer and using k-NN to search the reference data nearest to the embedded test samples. 

\subsubsection{Personalized Triplet Network (PTN)}
The role of the triplet neural network is to learn an embedding $f(\mathbf{x})$, for data $\mathbf{x}$ into a feature space $\mathbb{R}^d$ such that the euclidean distance between datum of the same target class ($y$) is small and the distance between datum of different target classes is large. With a squared Euclidean distance metric, triplet loss ($\mathcal{L}_T$) is defined by \citet{schroff_facenet:_2015} as:

\begin{equation}
\mathcal{L}_T = \sum_i^T \texttt{max} \Bigg\lbrace \bigg[ \| f(\mathbf{x}^a_i)-f(\mathbf{x}^p_i) \|^2_2 - \|f(\mathbf{x}^a_i) - f(\mathbf{x}^n_i) \| ^2_2 + \alpha \bigg] , 0 \Bigg\rbrace
\label{psm:eq:triploss}
\end{equation}

Where $\mathbf{x}^a_i$ is a sample from a given class (anchor), and $\mathbf{x}^p_i$ is a different sample of the same class (positive), and $\mathbf{x}^n_i$ is a sample of a different class (negative). $\alpha$ is the margin, which is a hyperparameter of the model defining the distance between class clusters. The same embedding $f(x)$ is applied to each sample in the triplet, and the objective is optimized over a training set of triplets with cardinality $T$. The number of possible triplets ($T$) that can be generated from a data set with cardinality $N$ is $\mathcal{O}(N^3)$.

\paragraph{}
In practice, TNNs converge well before a single pass over the full set of triplets \cite{schroff_facenet:_2015}, and therefore a subset of triplets must be specifically selected from the full set. We first implemented a naive strategy whereby triplets are randomly selected from $T$, enforcing only no temporal overlap between anchor and positive samples. We also implemented a triplet selection strategy where triplets derive their samples from a single subject, which yields a modified triplet loss function: 

\begin{equation}
\mathcal{L}_S = \sum_s^S \sum_i^{T_s} \texttt{max} \Bigg\lbrace \bigg[ \| f(\mathbf{x}^a_{s,i})-f(\mathbf{x}^p_{s,i}) \|^2_2 - \|f(\mathbf{x}^a_{s,i}) - f(\mathbf{x}^n_{s,i}) \| ^2_2 + \alpha \bigg] , 0 \Bigg\rbrace
\label{psm:eq:usertriploss}
\end{equation}

Where $\mathbf{x}^a_{s,i}$ is a segment of a particular activity class for subject $s$ (anchor), $\mathbf{x}^p_{s,i}$ a segment of the same activity class and subject of the anchor (positive), and $\mathbf{x}^a_{n,i}$ is a segment of a different activity class but from the same subject as the anchor (negative). $T_s$ denotes the full set of triplets that may be drawn from a single subject, and $S$ is the full set of subjects. This approach reduces the number of possible triplets to $\mathcal{O}(N)$. Various other strategies have been used in the computer vision domain to specifically select hard triplets for improving the efficiency of the TNN training \cite{schroff_facenet:_2015}.

We derive the PTN embedding $f(\mathbf{x})$ by training the FCN core with triplet loss. In our experiments, we evaluate conventional triplet loss with random triplets (PTN$^\dagger$ as per Eq. \ref{psm:eq:triploss}), and subject triplet loss (PTN as per Eq. \ref{psm:eq:usertriploss}) with a portion of the triplets being subject triplets and the remainder randomly selected. We used the same optimizer and hyperparameters as described previously in \ref{psm:sec:fcn}, except the learning rate was reduced to 0.0002 when training the FCN core with triplet loss. We used a value of 0.3 for the margin parameter $\alpha$. Despite the greater cardinality of the triplet set, we consistently define an epoch in this manuscript as having $N$ samples.

\subsection{Hardware and Software} 

The \texttt{keras} \cite{chollet_keras_2015} and \texttt{seglearn} \cite{burns_seglearn:_2018} open source python libraries were utilized to implement the machine learning models described in this work. The \texttt{scikit-learn} library was used to implement the k-nearest neighbor algorithm. Experiments were carried out locally on a computer with two NVIDIA Titan V GPUs for hardware acceleration. 

\section{Experiments} 

\subsection{Activity Classification} 

\begin{figure}[!ht]
  \begin{center}
    \includegraphics[scale=0.7]{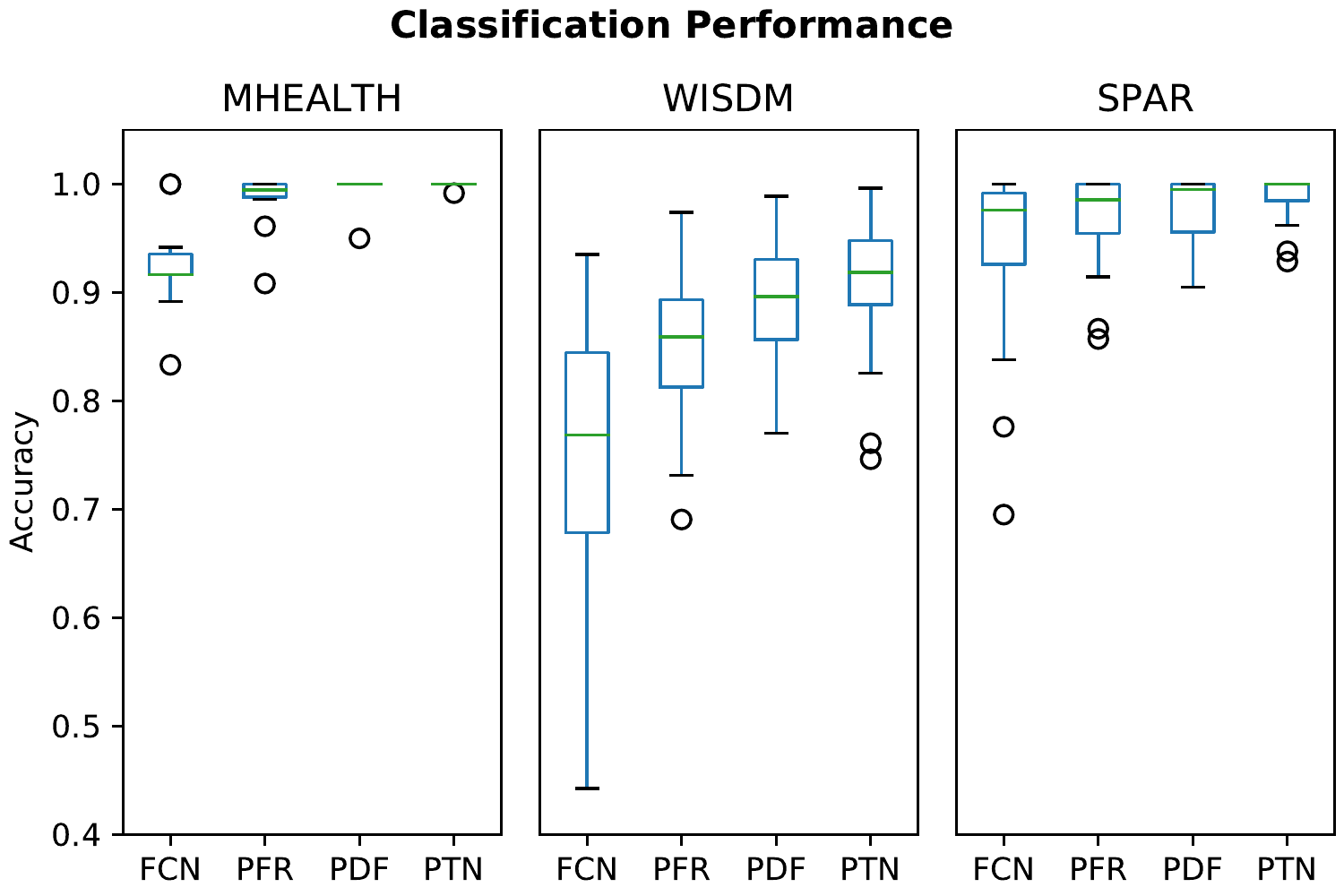}
  \end{center}
  \caption{Box and whisker plots showing distribution of classifier performance by subject using 5-fold cross validation. The personalized classifiers have better performance and less inter-subject performance variation than the impersonal FCN (fully convolutional network) model, with the PTN (personlized triplet network) model performing best overall.}
  \label{psm:f:baselines}
\end{figure}

\begin{figure}[]
  \begin{center}
    \includegraphics[scale=0.7]{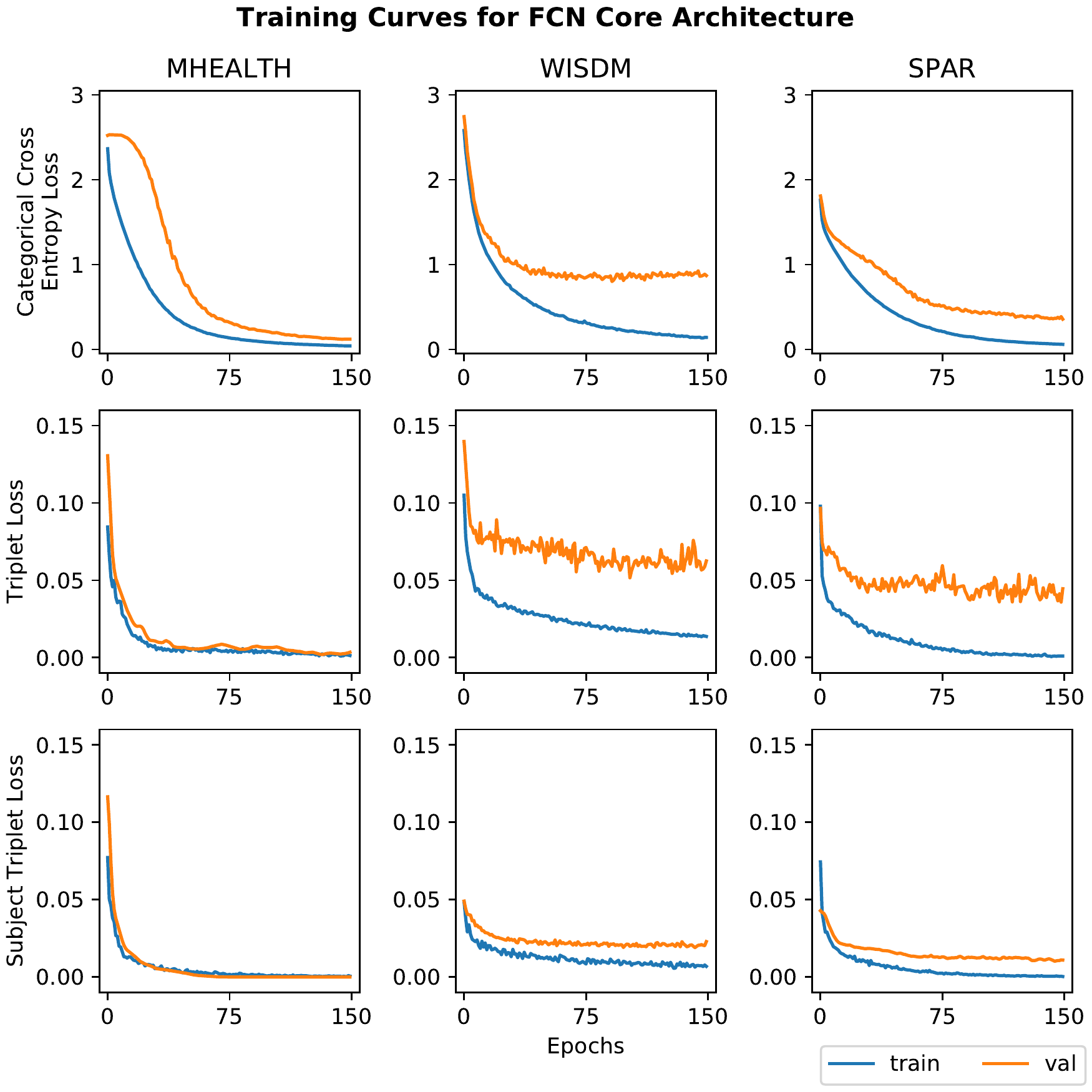}
  \end{center}
  \caption{Training and validation curves for FCN model (above), PTN model with conventional triplet loss (middle), and PTN model with subject triplet loss (bottom) over 150 epochs.}
  \label{psm:f:training}
\end{figure}

Classification accuracy was evaluated using 5-fold cross-validation grouping folds by subject. Subject distribution across folds was randomized but consistent for each algorithm in keeping with best practices for the evaluation of human activity recognition algorithms \cite{jordao_human_2019}. Cross-validated test set performance is summarized for each algorithm on the three data sets in Table \ref{psm:t:baseline}. Accuracy statistics (mean and standard deviation) are aggregated by subject, not by fold. Box and whisker plots demonstrating the variation in performance between individuals are provided in Figure \ref{psm:f:baselines}. Training curves for the FCN and PTN models are depicted in Figure \ref{psm:f:training}. 

\begin{table}[!ht]
\centering
\begin{tabular}{cccc}
\toprule
\textbf{Model} & \textbf{MHEALTH}  & \textbf{WISDM}    & \textbf{SPAR}     \\
\midrule
FCN	&	0.925 $\pm$ 0.049	&	0.754 $\pm$ 0.012	&	0.947 $\pm$ 0.069	\\
PEF	&	0.984 $\pm$ 0.029	&	0.852 $\pm$ 0.060	&	0.971 $\pm$ 0.038	\\
PDF	&	0.995 $\pm$ 0.016	&	0.889 $\pm$ 0.055	&	0.980 $\pm$ 0.028	\\
PTN$^\dagger$	&	0.993 $\pm$ 0.024	&	0.909 $\pm$ 0.054	&	0.978 $\pm$ 0.035	\\
PTN	&	\textbf{0.999} $\pm$ \textbf{0.003}	&	\textbf{0.913} $\pm$ \textbf{0.053}	&	\textbf{0.990 $\pm$ 0.017}	\\

\bottomrule \\
\end{tabular}
\caption{Cross-validated classification accuracy (mean $\pm$ standard deviation) aggregated by subject. PTN$^\dagger$ was trained with conventional triplet loss, and PTN was trained with 50\% subject triplets. FCN: Fully convolutional network (impersonal), PEF: personalized engineered features, PDF: personalized deep features, PTN: personalized triplet network.}
\label{psm:t:baseline}
\end{table}

\paragraph{}
We found that all the personalized feature classifiers out-performed the impersonal FCN classifier and reduced the incidence and degree of negative outlier subjects who individually experience poor performance in the impersonal model. This effect is seen in Figure \ref{psm:f:baselines}. Both the personalized deep feature models (PDF and PTN) outperformed the personalized engineered features (PEF). Specifically, the PTN model utilizing subject triplet loss had the best classification performance. 

\subsection{Out-of-Distribution Detection}

We assessed model performance for distinguishing activity classes present in the training distribution from unknown (out-of-distribution) activity classes. This evaluation was performed by training the models on a subset (70\%) of the activity classes, and testing with the full set of activity classes in a subject group 5-fold cross validation scheme. In each fold, the classes considered out-of-distribution were randomly selected but were consistent across the algorithms evaluated. Out-of-distribution performance was assessed using the area under the receiver operating curve (AUROC) for the binary classification task of in- vs out-of-distribution. 

\begin{figure}[!ht]
  \begin{center}
    \includegraphics[scale=0.7]{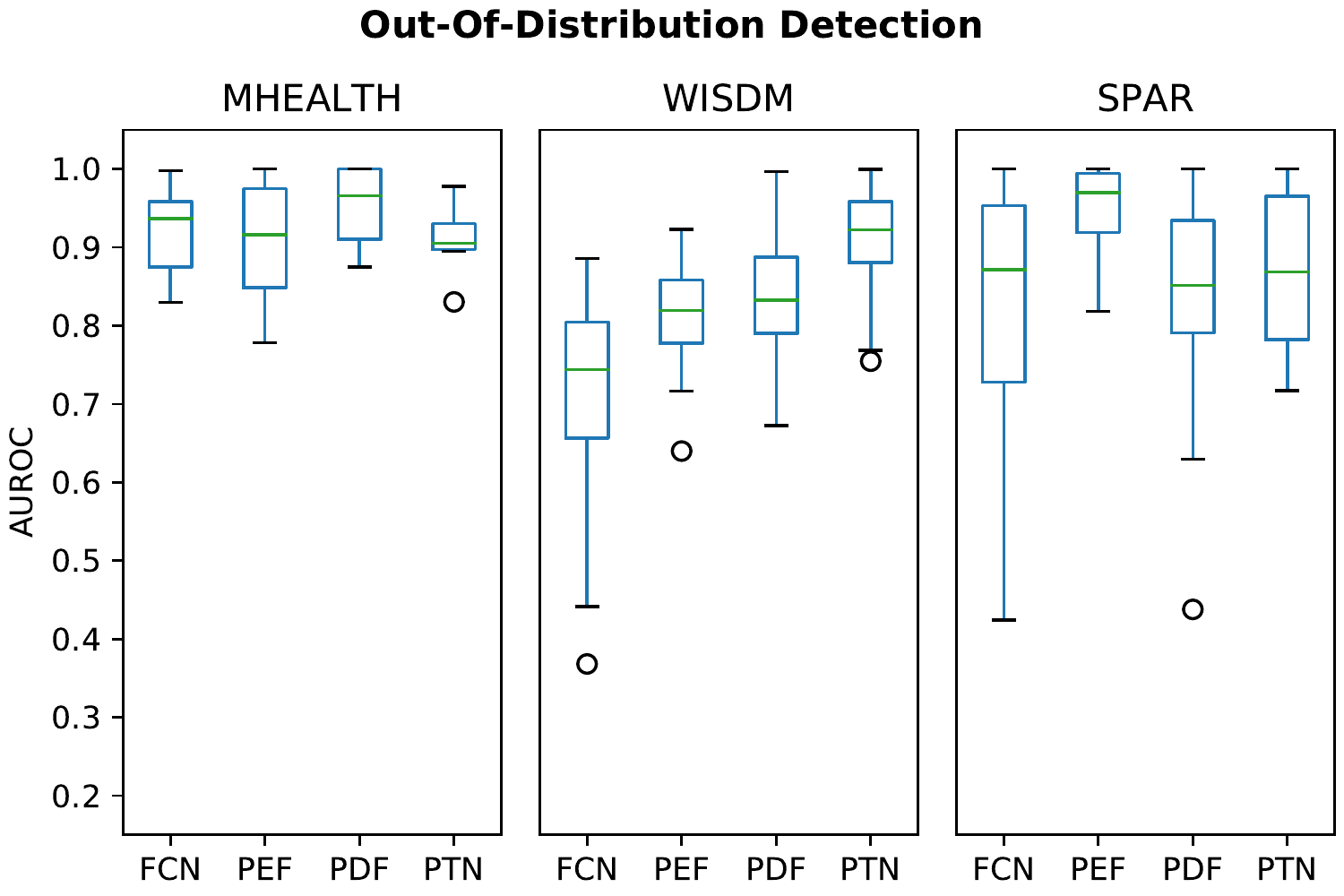}
  \end{center}
  \caption{Box and whisker plots showing distribution of OOD detection performance across subjects, with 30\% of activity classes held back from the training set. The PDF, PTN, and PEF classifiers had the highest mean OOD scores for the MHEALTH, WISDM, and SPAR data sets respectively.}
  \label{psm:f:ood}
\end{figure}

Out-of-distribution (OOD) classification was implemented for the personalized feature classifiers by thresholding mean euclidean distance from  the three nearest neighbors in the embedded reference data. For the FCN model, we used the maximum softmax layer output as the decision function. OOD detection performance is plotted in Figure \ref{psm:f:ood}. In contrast to the classification task, the best performing OOD detector appeared to depend on the data set tested. The PDF, PTN, and PEF classifiers had the highest mean OOD scores for the MHEALTH, WISDM, and SPAR data sets respectively. The PTN model performed best overall for OOD detection with AUROC scores near 90\% across all datasets.

\subsection{Generalization to New Activity Classes}

\begin{figure}[!ht]
  \begin{center}
    \includegraphics[scale=0.7]{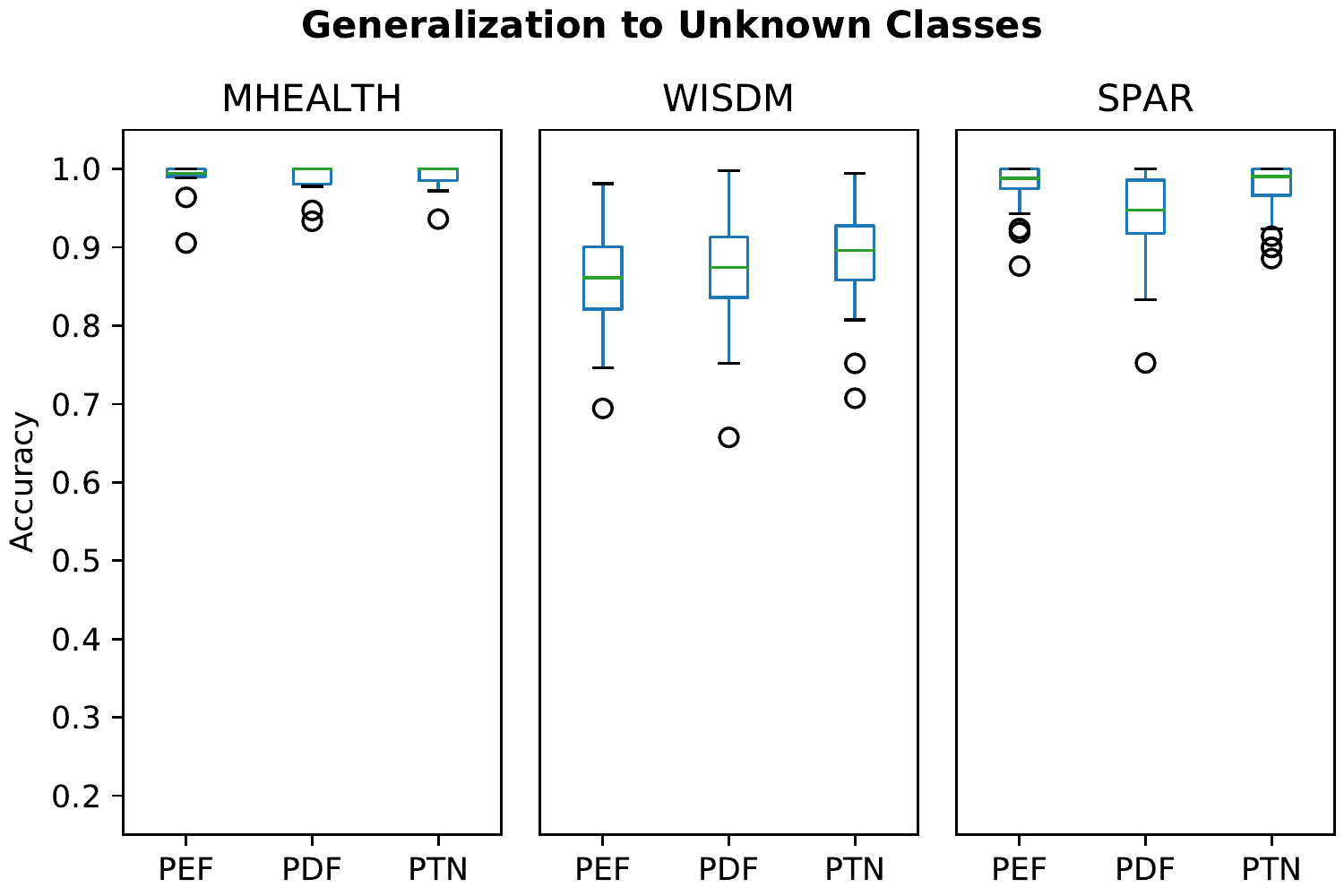}
  \end{center}
  \caption{Distribution of activity classification performance when generalizing an embedding to novel activity classes, with 30\% of activity classes held back from the training set. The PTN model had the best mean performance across all three data sets.}
  \label{psm:f:generalize}
\end{figure}

Generalization of the personalized features to new activity classes was assessed in a manner similar to the out-of-distribution detection. Instead of a binary in- vs out- classification target, accuracy for classification was assessed across the full distribution of activities in the context of training the embedding on the reduced (in-distribution) subset of activities. The FCN model was not assessed for this task as generalization to new target classes is not a feature of the softmax classification layer. The results are plotted in Figure \ref{psm:f:generalize}, depicting excellent performance of all three feature classifiers with the PTN algorithm performing best overall across all three data sets. 

\subsection{Reference Data Size}

\begin{figure}[!ht]
  \begin{center}
    \includegraphics[scale=0.7]{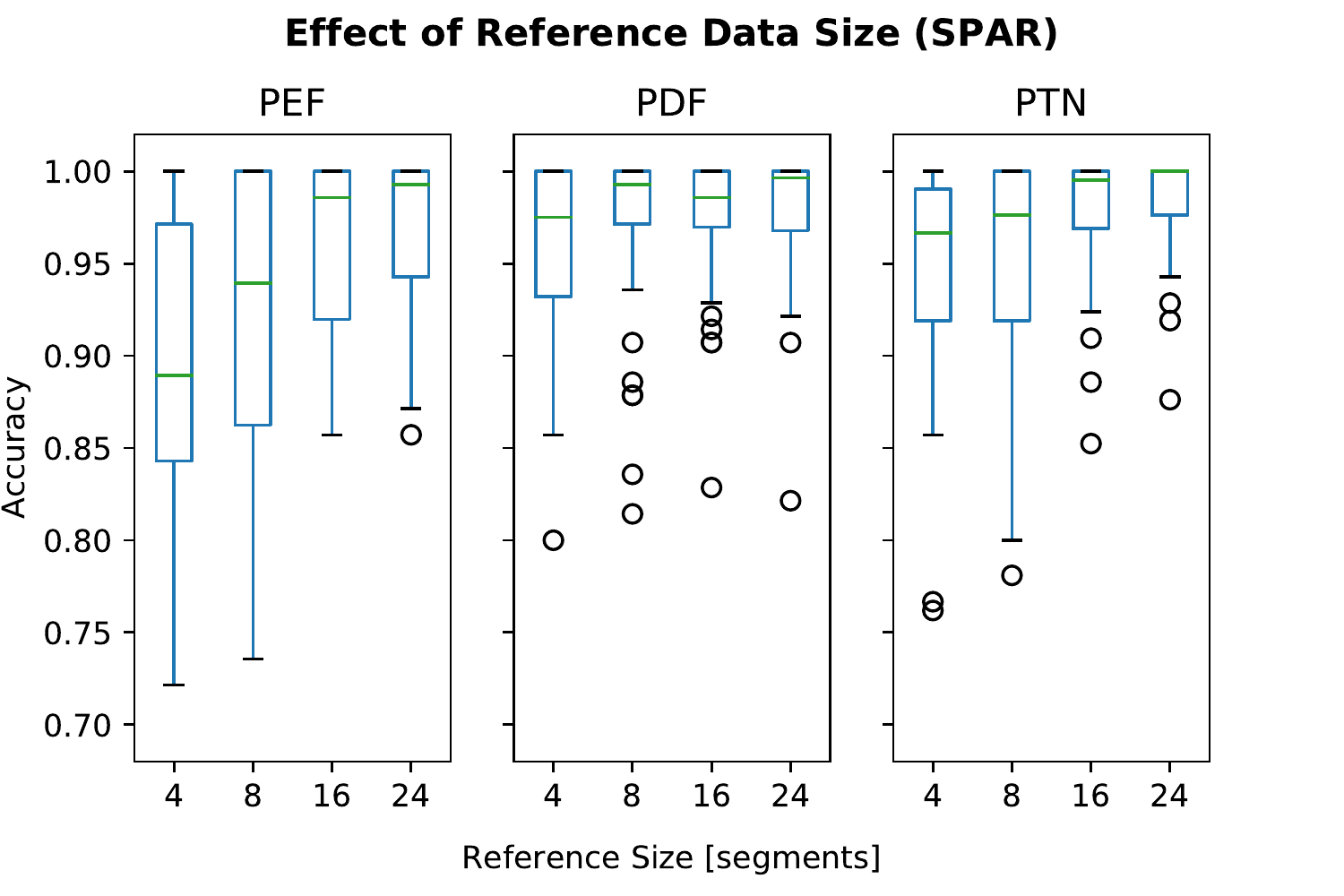}
  \end{center}
  \caption{The effect of reference data size (number of reference segments per activity class) on personalized feature classifier accuracy, evaluated on the SPAR data set. Increasing reference data size improves performance for all personalized algorithms.}
  \label{psm:f:support}
\end{figure}

The effect of reference sample quantity on personalized feature classifier accuracy was evaluated using 5-fold cross validation on the SPAR data set. The results are plotted in Figure \ref{psm:f:support}. We found approximately 16 reference segments (equal to approximately 16 seconds of data) are required per activity class to achieve optimal results.

\subsection{Embedding Size}
A parsimonious representation of activity is desirable to minimize the storage and computational cost of personalized feature inference. We assess the effect of embedding size on model performance using 5-fold cross validation on the SPAR data set. For the PDF and PTN models, the embedding size was adjusted at the final dense layer of the FCN core. For the engineered features, we reduced the embedding size by selecting the most important features as ranked using Gini importance \cite{louppe_understanding_2013}. The Gini importance was calculated for the engineered features using an Extremely Randomized Trees classifier \cite{geurts_extremely_2006} with an ensemble of 250 trees.

\begin{figure}[!ht]
  \begin{center}
    \includegraphics[scale=0.7]{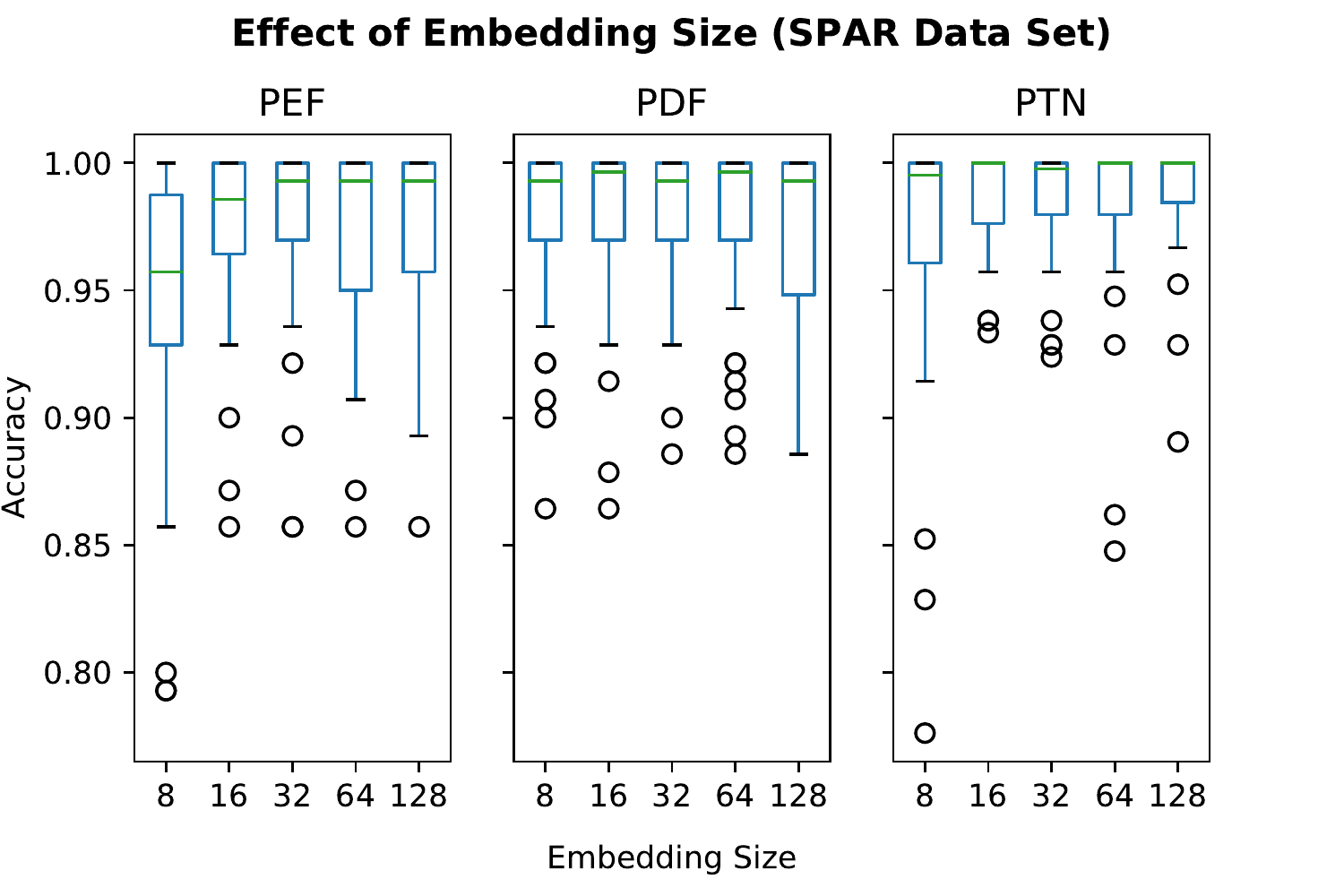}
  \end{center}
  \caption{The effect of embedding size (number of features) on personalized feature classifier accuracy, evaluated on the SPAR data set. This demonstrates how both the PDF and PTN models can maintain high performance with a parsimonious embedding, whereas the PEF model performance is significantly degraded.}
  \label{psm:f:embedding}
\end{figure}

Classifier performance as a function of embedding size is plotted in Figure \ref{psm:f:embedding}. We found that accurate results are achievable down to an embedding size of 8, and that the PDF and PTN models are superior to PEF for learning a parsimonious embedding. 

\subsection{Computational Expense}

The computational cost for each model on the SPAR data set is reported in Table \ref{psm:t:cost}, detailing training and inference time on our hardware, and storage size for model and reference data. In our implementation, the inference time for the PDF and PTN classifiers was split nearly equally between embedding computation and nearest embedding search. Training the FCN core with triplet loss in the PTN model increased the fit time by approximately 5-fold in comparison to training with categorical cross entropy loss as done for the PDF and FCN models. 

\begin{table}[]
\centering
\begin{tabular}{ccccc}
\toprule
\textbf{Model} & \textbf{Fit Time {[}s{]}} & \textbf{Inference Time {[}s{]}} & \textbf{Model Size {[}kB{]}} & \textbf{Reference Size {[}kB{]}} \\ \midrule
FCN	&	137	&	0.47	&	4290	&	0	\\
PEF	&	3.3	&	0.39	&	3.8	&	112	\\
PDF	&	129	&	0.94	&	1095	&	112	\\
PTN	&	667	&	1.3	&	1095	&	112	\\
\bottomrule \\
\end{tabular}
\caption{Computational and storage expense for SPAR data set on a single fold (test size 0.2). Reference size for personalized feature classifiers is based on single precision 64 feature embeddings, with 16 samples for each of the 7 activity classes.}
\label{psm:t:cost} 
\end{table}

\section{Discussion}
This work described and evaluated two novel methods for personalized inertial human activities using personalized deep features (PDF) and a personalized triplet network (PTN) and compared these to a baseline impersonal fully convolutional network (FCN) and personalized engineered features (PEF) for inertial human activity recognition. 

\paragraph{}
The PTN and PDF models significantly outperformed PEF for personalized activity recognition across all metrics evaluated. The three personalized feature classifiers also significantly outperformed the impersonal FCN classifier, which represents current state of the art for the field. In fact, the personalized classifiers were able to achieve performance approaching training set performance of the impersonal FCN classifier. As predicted, we found that the FCN classifier, while producing excellent results over all, performed poorly for some individuals (as low as 50\% accuracy) as shown in Figure \ref{psm:f:baselines}. The three personalized feature classifiers evaluated all significantly mitigated this issue, and had more consistent results across individual subjects in each data set, a chief objective of this work.

\paragraph{}
The PTN model had the best performance overall, exceeding that of the personalized deep features learned by the PDF model serendipitously in a conventional supervised learning framework. However, a significant disadvantage of using a triplet neural network to learn the embedding function is the increased computational cost during training. On our hardware, the PTN approach increases the training time five fold and triples the GPU memory requirements in comparison to training an identical core with categorical cross entropy loss. This is because there is the further cost of triplet selection and each triplet is comprised of three distinct samples that must each be embedded to compute the triplet loss.  Fortunately, once the embedding has been trained, there is little difference in computational requirements to compute the embedding or classify an unknown sample. 

\paragraph{}
A significant advantage of using personalized features for activity classification is the built in capability to use them for out-of-distribution activity detection and classification of novel activities. In the human activity recognition field, out-of-distribution detection is particularly important as there exists an infinite number of possible human actions, and therefore it may be impractical to include all possible actions in the training set or even all reasonably likely actions. Typically, it is a desirable property of a HAR system that it can be trained to recognize a select number of pertinent activities and have the ability to reject anomalous activities not in the training-distribution. 

\paragraph{}  
Typically, deep learning classification algorithms implementing a soft-max output layer perform poorly at out-of-distribution activity detection due to overconfidence \cite{guo_calibration_2017}. This well-known result was replicated in our evaluations of the FCN classifier on the WISDM and SPAR data sets. Interestingly, the FCN classifier had good OOD performance on MHEALTH. That said, various approaches to improving OOD performance for neural networks have been investigated in the computer vision fields with mixed results and this remains an active area of research \cite{geng_recent_2019}. Yet, there has been relatively little work on this problem in the context of time series data and particularly inertial activity recognition \cite{casale_personalization_2012, yang_open-set_2019, roitberg_informed_2018}. We have demonstrated that mean nearest neighbor distance with personalized features has good performance for our synthetic OOD evaluation. However, further work is required to evaluate alternative approaches and build true out-of-distribution data sets incorporating real-world variation in subject daily activities.

\paragraph{}
The personalized k-nn model used to search reference embeddings for classification of test samples in the PEF, PDF, and PTN models was found to be effective, but approximately doubles the inference time in comparison to the FCN model that used a softmax layer for classification. A disadvantage with k-nn search is that computational time complexity and data storage requirement scales with the number of reference samples $\mathcal{O}(N)$. This property of k-nn limits its utility as an impersonal classifier, as performing inference requires searching the entire training data set. In the context of a personalized algorithm, however, the k-nn search is limited only to the subject's reference samples, which we have demonstrated need only include tens of samples per activity class. Of course, other search strategies could be implemented to search the reference data. The nearest centroid method, for instance, could be used which has computational complexity $\mathcal{O}(1)$, scaling linearly with number of reference classes.

\paragraph{}
The FCN core architecture described in this work, with just 278,848 parameters ($\sim$1 MB), is a relatively compact model. Particularly, in comparison to computer vision or language models that can exceed tens or hundreds of millions of parameters \cite{szegedy_inception-v4_2016, he_deep_2015, vaswani_attention_2017}. Given the small size of the model and reference embeddings, implementing a personalized feature classifier based on the FCN core may be feasible within an edge computing system where the computations for activity recognition are performed locally on the user's hardware (mobile device). There are various advantages of an edge computing approach, including improved classification latency, reliability, and network bandwidth usage \cite{ajerla_real-time_2019}. 

\paragraph{}
The novel triplet loss function (Eq. \ref{psm:eq:usertriploss}) and triplet selection strategy described in this work significantly improved the performance of the PTN model in comparison to conventional triplet loss. The subject triplets can be considered "hard" triplets in the context of other strategies for specifically selecting hard triplets to improve TNN training \cite{schroff_facenet:_2015, wang_learning_2014, wang_learning_2016, yu_correcting_2018}. How well our approach compares to other hard triplet selection strategies remains as future work. However, our strategy may be worth considering as it is straight forward to implement and computationally inexpensive in comparison to strategies that require embeddings to be computed prior to triplet selection. The benefit of subject triplets may hold to a greater extent on data sets collected with heterogenous hardware. Certainly, our work demonstrates that triplet selection is essential to achieving good performance with TNNs in the inertial activity recognition context. 

\paragraph{}
Although there was no temporal overlap in the segments used to derive the reference and test embeddings, it is a limitation of this work that they were derived from the same time series. Unfortunately, we are not a aware of any inertial activity recognition data sets that contain repeated data collections of the same activity classes by the subjects. Certainly, such a data set would be worthwhile to collect and would serve as the best validation of the approaches described in this work. 

\section{Conclusion}
Deep embeddings derived from fully convolutional neural networks trained with categorical cross entropy or triplet loss significantly outperform engineered features for personalized activity recognition and outperform impersonal supervised learning approaches. Subject triplet selection improves the training and performance of triplet neural networks used for personalized activity recognition. 

\bibliography{references}  

\end{document}